\documentclass[sigconf]{acmart}

\copyrightyear{2026}
\acmYear{2026}
\setcopyright{cc}
\setcctype{by}
\acmConference[MM '26]{Proceedings of the 34th ACM International Conference on Multimedia}{November 10--14, 2026}{Rio de Janeiro, Brazil}
\acmBooktitle{Proceedings of the 34th ACM International Conference on Multimedia (MM '26), November 10--14, 2026, Rio de Janeiro, Brazil}
\acmDOI{10.1145/3767308.3838688}
\acmISBN{979-8-4007-2213-4/2026/11}

\begin{document}

\title[Generalization over Memorization]{Generalization over Memorization: \\Generalization-Aware Diffusion Adaptation for Single-Image Multi-View Synthesis}

\author{Jie Li}
\authornote{These authors contributed equally to this work.}
\orcid{0009-0009-2509-9071}
\affiliation{%
  \institution{Huazhong University of Science and Technology}
  \city{Wuhan}
  \country{China}}
\email{364670263@qq.com}

\author{Xingchen Zou}
\authornotemark[1]
\orcid{0009-0004-4362-6617}
\affiliation{%
  \institution{The Hong Kong University of Science and Technology (Guangzhou)}
  \city{Guangzhou}
  \country{China}}
\email{xzou428@connect.hkust-gz.edu.cn}

\author{Yuxuan Liang}
\orcid{0000-0003-2817-7337}
\authornote{Corresponding author.}
\affiliation{%
  \institution{The Hong Kong University of Science and Technology (Guangzhou)}
  \city{Guangzhou}
  \country{China}}
\email{yuxuanliang@hkust-gz.edu.cn}

\begin{abstract}
We present the winning solution to the ACM Multimedia 2026 Grand Challenge on Single-Image Guided Multi-Angle Image Synthesis. It ranks first among 293 registered teams; 56 teams obtained at least one scored submission on the public Phase-A leaderboard. With only 40 training scenes, the challenge requires 26 target views from one RGB model and one forward pass per view; it prohibits explicit geometry, external rendering, chained generation, candidate selection, and post-processing. We identify a critical model-selection failure: shared training and validation scenes make memorization appear as transferable view control. We therefore introduce GoM. Short for \emph{Generalization over Memorization}, the framework combines scene-disjoint validation, exposure-matched selection, and targeted diffusion adaptation. Its synthesis model adapts a 4B rectified-flow DiT using rank-32 LoRA, optimizer restarts, late-checkpoint averaging, and VAE decoder tuning. More than 300 offline experiments and 24 online submissions show that validation design and training-trajectory control can matter as much as architecture scale in small-data generative modeling.
\end{abstract}



\maketitle
\hypersetup{%
  pdfauthor={Jie Li, Xingchen Zou, Yuxuan Liang},
  pdfkeywords={single-image novel view synthesis; diffusion models; parameter-efficient fine-tuning; generalization}}

\section{Introduction}
Novel-view synthesis turns one reference image into observations from new camera viewpoints, supporting 3D asset creation, virtual production, content editing, and immersive visualization. Achieving this capability from a single image is fundamentally ambiguous: geometry, appearance, and occluded regions must be inferred from incomplete evidence while the identity of the original scene remains intact. A successful system must therefore balance plausible generation with faithful view control.

Large diffusion models have advanced this problem by providing strong visual and geometric priors. Zero-1-to-3 learns camera control from synthetically rendered object views~\cite{liu2023zero123}; ZeroNVS extends zero-shot synthesis to complex scenes through 6-DoF conditioning and diverse scene data~\cite{sargent2024zeronvs}; and CAT3D combines joint multi-view diffusion with robust 3D reconstruction~\cite{gao2024cat3d}. Although effective, these systems rely on camera parameters, pose-annotated multi-view training, explicit geometry, or multi-stage reconstruction. This leaves an important gap: reliable view transformation from few RGB scenes within one end-to-end image model.

The Single-Image Guided Multi-Angle Image Synthesis Grand Challenge at ACM Multimedia 2026 provides a controlled setting for studying this gap. Given one $1280\times720$ RGB photograph, a system must synthesize the same scene from 26 directions, elevations, and viewing distances. The rules permit only one model and one forward pass per view, excluding intermediate geometry, external rendering, chained generation, candidate selection, and post-processing. The official training set contains only 40 scenes under strict compute budgets, while evaluation jointly measures pixel, structural, perceptual, and no-reference quality (Sec.~\ref{sec:setup}). Phase A gives development feedback on targets from training-overlapping scenes; Phase B uses unseen scenes to determine the final ranking. The track registered 293 teams, of which 56 produced at least one scored Phase-A submission; our solution won the final ranking. Its development raises two coupled research questions: how should an end-to-end model be adapted under severe data constraints, and how can genuine progress be distinguished from memorization?

\begin{figure*}[t]
  \centering
  \includegraphics[width=\textwidth]{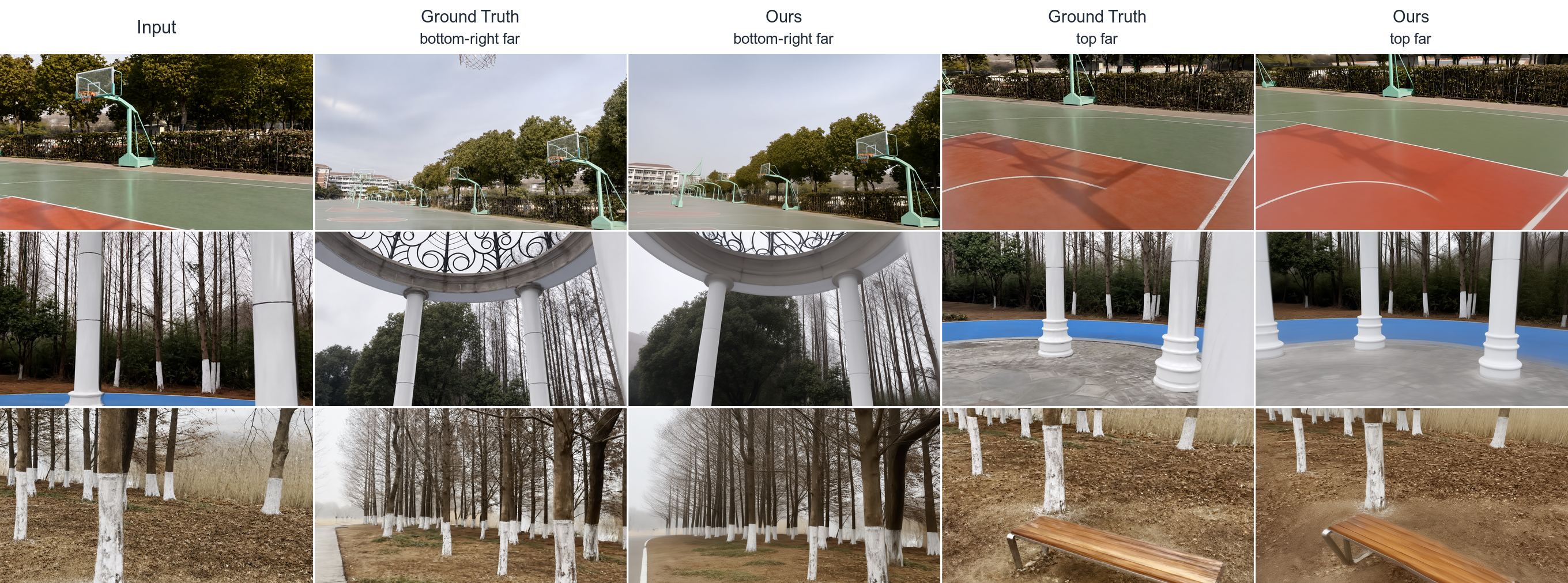}
  \caption{Qualitative results. From a \emph{center-medium} input, B2 synthesizes \emph{bottom-right-far} and \emph{top-far} views; predictions are paired with ground truth across three scenes.}
  \Description{Three rows of qualitative results. Each row shows one input followed by ground-truth and model outputs for bottom-right-far and top-far target views. The scenes depict a basketball court, a pavilion, and a forest.}
  \label{fig:qualitative}
\end{figure*}

The second question is easily overlooked. Although Phase A is useful for development, its scene overlap can reward memorization. Forty scenes yield 1{,}040 pairs, but view-level splitting retains scene identity because many targets share a source. A model may therefore improve by recovering known-scene appearance rather than transferable view control. Consistent with this diagnosis, the same container drops from 1.9386 in Phase A to 1.8512 in Phase B, with most of the decline arising from structural and perceptual consistency rather than pixel-level color fidelity. Model selection is therefore not merely an implementation detail; it is part of the generalization problem.

We address both questions with \emph{Generalization over Memorization} (GoM). Its selection layer uses scene-disjoint validation and exposure-matched transfer estimates to distinguish transferable gains from scene recall. Its synthesis layer combines rank-32 LoRA on a 4B rectified-flow DiT, optimizer restarts, checkpoint averaging, and VAE-decoder adaptation (Sec.~\ref{sec:methodology}). Across 24 Phase-A submissions and more than 300 offline experiments, GoM retains robust interventions while rejecting gains confined to overlapped scenes (Sec.~\ref{sec:falsification}).

\begingroup
\setlength{\parskip}{0pt}
\setlength{\listisep}{2pt}
Our contributions are:
\begin{itemize}
\setlength{\topsep}{2pt}
\setlength{\partopsep}{0pt}
\setlength{\itemsep}{0pt}
\setlength{\parsep}{0pt}
\item \textbf{Winning solution.} We develop the 1st-place solution to the ACM MM 2026 Single-Image Guided Multi-Angle Image Synthesis Grand Challenge. Among 293 registered teams, 56 obtained a scored Phase-A submission; GoM won under a single-model, one-pass-per-view protocol without geometry or post-processing. On unseen Phase-B scenes, it leads SSIM, NRIQA, and the composite score.
\item \textbf{Generalization-first methodology.} We introduce GoM to separate scene memorization from transferable view control through scene-disjoint validation, exposure matching, and transfer estimation. It identifies a compact recipe combining LoRA, trajectory regularization, and decoder adaptation.
\item \textbf{Systematic empirical study.} Across more than 300 offline experiments and 24 online submissions, we audit over 40 unsuccessful directions, quantify offline-to-online transfer, and derive practical lessons for small-data generative modeling. The audit yields reusable decision rules rather than a leaderboard-only report.
\end{itemize}
\endgroup

\section{Related Work}
\paragraph{Single-image novel-view synthesis.}
Zero-1-to-3~\cite{liu2023zero123} learns camera control from synthetic object views, while ZeroNVS~\cite{sargent2024zeronvs} extends zero-shot synthesis to complex scenes through 6-DoF conditioning and diverse scene data. SV3D~\cite{voleti2024sv3d} adapts latent video diffusion with explicit camera control to generate consistent orbital views, and CAT3D~\cite{gao2024cat3d} jointly generates pose-conditioned views for robust 3D reconstruction. For scene-level synthesis, MultiDiff~\cite{muller2024multidiff} combines monocular depth, reference-view warping, and a video-diffusion prior to generate an entire view sequence jointly. These systems demonstrate the strength of geometric and temporal priors, but their camera parameters, depth/warping intermediates, or joint multi-frame generation fall outside the challenge protocol. Our contribution instead concerns reliable selection and adaptation of a single RGB image editor~\cite{brooks2023instructpix2pix} driven only by a view-name instruction.

\paragraph{Efficient adaptation and trajectory averaging.}
LoRA~\cite{hu2022lora} adapts large pretrained models through low-rank parameter updates. Stochastic weight averaging~\cite{izmailov2018swa}, model soups~\cite{wortsman2022soups}, and snapshot ensembles~\cite{huang2017snapshot} improve robustness by combining states along one or more training trajectories; SGDR~\cite{loshchilov2017sgdr} additionally couples restarts with learning-rate annealing. GoM draws on these ideas but studies them in a small-data diffusion regime where optimization gains can be confounded by scene memorization.

\paragraph{Latent reconstruction bottlenecks.}
Latent diffusion models~\cite{rombach2022ldm} synthesize through a VAE whose decoder bounds pixel fidelity. Parameter-efficient tuning typically adapts the denoiser while leaving this reconstruction path fixed. We therefore complement DiT LoRA with decoder adaptation, optimizing generation and reconstruction bottlenecks separately.

\section{Methodology}
\label{sec:methodology}

\begin{figure*}[t]
  \centering
  \includegraphics[width=0.99\textwidth]{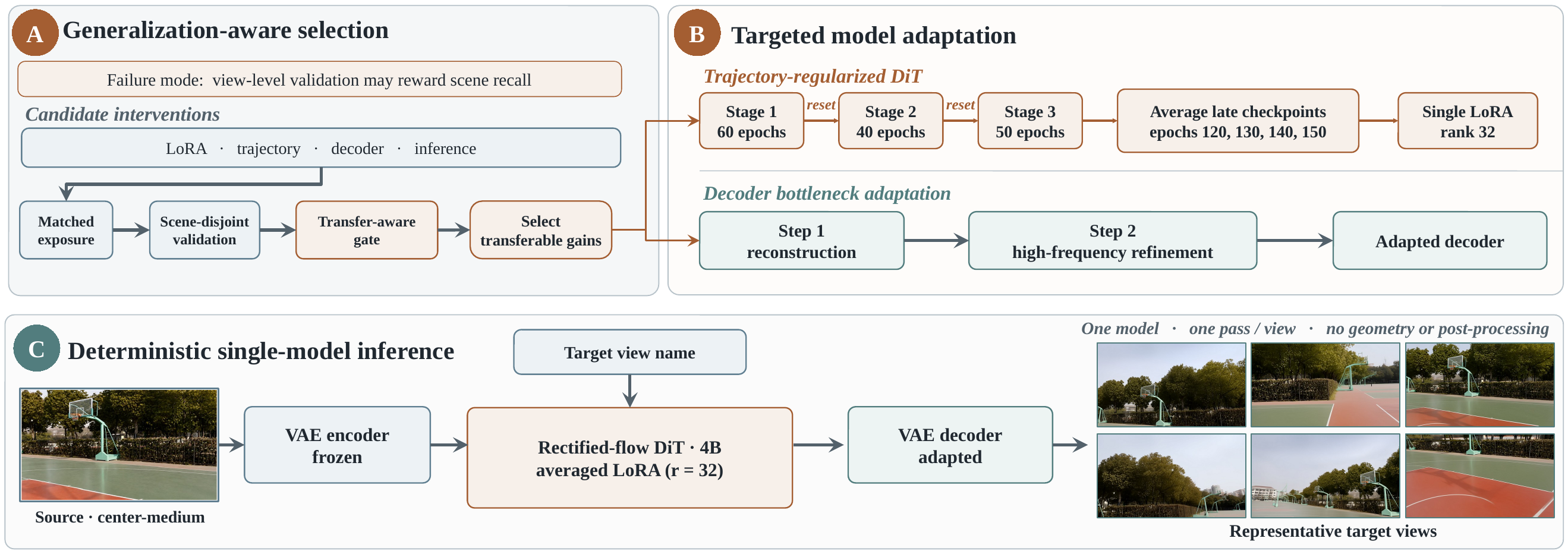}
  \vspace{-1em}
  \caption{Overview of GoM: (A) transfer-aware model selection, (B) staged LoRA and decoder adaptation, and (C) single-model inference from one source to requested views without geometry or post-processing.}
  \Description{A three-panel overview of Generalization over Memorization. Panel A shows generalization-aware model selection. Panel B shows staged LoRA training with checkpoint averaging and two-step decoder adaptation. Panel C uses real experimental images to show deterministic inference from one center-medium source image to six representative requested views.}
  \label{fig:pipeline}
\end{figure*}

\subsection{Overview}
GoM addresses two coupled bottlenecks exposed in Sec.~1. First, scene-overlapping validation can reward memorization and mislead model selection. Second, the strict end-to-end setting requires useful view control without explicit geometry. We therefore organize GoM into a \emph{model-selection layer}, which decides which gains are likely to transfer, and a \emph{synthesis layer}, which adapts a pretrained diffusion model under the challenge constraints. The former provides trustworthy evidence for the latter; Figure~\ref{fig:pipeline} summarizes the resulting synthesis pipeline.

\subsection{Generalization-Aware Model Selection}
\paragraph{Selection objective.}
Let $b$ denote a matched baseline and $c$ a candidate intervention. For evaluation regime $s$, GoM records the paired improvement
\[
\Delta_s(c;b)=S_s(c)-S_s(b), \qquad s\in\{\mathrm{audit},\mathrm{sd}\},
\]
where ``audit'' denotes the early overlapping-scene diagnostic and ``sd'' the scene-disjoint sister lineage. For an isolated intervention axis $a$, paired Phase-A submissions calibrate $r_a=\Delta_A/\Delta_{\mathrm{audit}}$. Final selection then estimates $\widehat{\Delta}_A(c)=r_a\Delta_{\mathrm{sd}}(c;b)$. A candidate advances only if its predicted gain exceeds 0.005, lies outside the observed training-variation band, and respects the challenge's compute and component-quality constraints. Composite packages without an isolated $r_a$ require direct scene-disjoint evidence. The objective is therefore not to maximize an overlapping validation score, but to maximize the gain expected to survive a change of scenes.

\paragraph{Scene-disjoint validation.}
The main motivation is to make validation measure transferable view control rather than scene recall. GoM therefore treats scene identity, not target view, as the unit of generalization. Development uses two distinct regimes. An early six-scene audit contains 156 target views, but its LoRA lineage was trained on all 40 scenes; we consequently use it only for within-lineage deltas and never interpret its absolute score as unseen-scene performance. For final B2 selection, we train a 34-scene sister lineage that excludes all six audit scenes from both DiT and decoder optimization. The averaging window, inference protocol, and decoder refinement are selected in this scene-disjoint regime before the chosen recipe is retrained on all 40 scenes for submission. This separation prevents an overlapping-scene diagnostic from being mistaken for evidence of generalization.

\paragraph{Transfer prioritization.}
The limited submission budget makes exhaustive online evaluation impossible. Table~\ref{tab:transfer} reports the axes for which both offline and online deltas are isolated. High-transfer axes receive priority, whereas composite updates and gains confined to one metric or one scene are treated cautiously rather than assigned a spurious single-axis ratio.

\begin{table}[t]
  \caption{Offline-to-online transfer calibration from paired comparisons.}
  \label{tab:transfer}
  \small
  \setlength{\tabcolsep}{4pt}
  \begin{tabular}{lrrr}
    \toprule
    \textbf{Intervention axis} & \textbf{Offline $\Delta$} & \textbf{Online $\Delta$} & \textbf{$r_a$} \\
    \midrule
    Checkpoint averaging & $+0.0202$ & $+0.0169$ & \textbf{84.0\%} \\
    Decoder adaptation & $+0.0114$ & $+0.0089$ & 78.1\% \\
    Seed selection & $+0.0170$ & $+0.0092$ & 54.1\% \\
    Adam restarts & $+0.0483$ & $+0.0109$ & 22.6\% \\
    \bottomrule
  \end{tabular}
\end{table}

\paragraph{Controlled comparisons.}
To keep these estimates interpretable, each submission is associated with one primary hypothesis. Alternatives are compared at matched training exposure, and absolute validation scores are not compared across unrelated lineages. Repeated DDP runs vary by approximately $\pm0.006$, so gains within this band are treated as indistinguishable. An offline result receives a leaderboard submission only when its transfer-adjusted gain exceeds $0.005$. These rules turn the leaderboard from a tuning target into a sparse source of generalization evidence.

\subsection{End-to-End Synthesis Model}
The challenge rules motivate a geometry-free image-editing formulation. The model builds on FLUX.2-klein~\cite{flux2}, a 4B-parameter rectified-flow DiT~\cite{lipman2023flow,liu2023rectified,esser2024scaling}. The VAE encoder and text encoder remain frozen, the DiT is adapted with LoRA~\cite{hu2022lora}, and the VAE decoder is fine-tuned separately. For each requested view, a deterministic chain maps the input through the VAE encoder, LoRA-adapted DiT, and VAE decoder. The view is specified only by its name (e.g., ``left near''), avoiding camera parameters and intermediate geometry.

The official data form 1{,}040 triplets from 40 scenes and 26 targets per scene. Each triplet contains the shared \emph{center-medium} input, target-view text, and target RGB image. The final recipe samples every pair twice per epoch and uses no geometric augmentation.

\subsection{Trajectory-Regularized DiT Adaptation}
With only 1{,}040 training pairs, continuously optimizing the same objective can convert late validation gains into scene memorization. GoM therefore regularizes the training trajectory rather than enlarging the backbone. We insert rank-32 LoRA adapters into all attention projections and the output projections of the 20 single-transformer blocks. Training uses bf16, four-GPU DDP, gradient checkpointing, and a constant learning rate of $10^{-4}$. Instead of one continuous run, GoM divides 150 epochs into stages of 60, 40, and 50 epochs. Each stage resumes the learned weights but reinitializes the Adam moments, perturbing the trajectory without changing the learning-rate schedule.

The final stage saves a checkpoint every five epochs. We average selected checkpoints tensor by tensor into one LoRA file, preserving the single-model submission requirement. B1 averages epochs $\{130,140,150\}$, while the scene-disjoint sweep for B2 selects $\{120,130,140,150\}$. The role of the restarts and averaging window is evaluated in Sec.~\ref{sec:results}.

\subsection{Decoder Bottleneck Adaptation}
\label{sec:decoder}
LoRA leaves the VAE reconstruction path unchanged: its 32.4\,dB domain reconstruction exposes a separate fidelity bottleneck. We therefore adapt only the 49.6M-parameter decoder, keeping the encoder and DiT fixed.

\paragraph{Step 1: reconstruction-only adaptation.}
We optimize $L_1+0.5\,\mathrm{MSE}$ for 1{,}800 steps at learning rate $8\times10^{-6}$. B1 excludes the six validation scenes from this stage; after the configuration is fixed, B2 retrains the same objective on all 40 scenes.

\paragraph{Step 2 (B2 only): high-frequency refinement.}
To recover fine texture, B2 continues for 1{,}800 steps on 11{,}273 unlabeled natural images from KonIQ-10k~\cite{hosu2020koniq} and UHD-IQA~\cite{hosu2024uhdiqa}:
\[
L_1+0.5\,\mathrm{MSE}+0.5\,L_1\!\big(\nabla^2 \hat{x},\,\nabla^2 x\big)+0.02\,\big(1-\mathrm{IQA}(\hat{x})\big),
\]
where $\nabla^2$ denotes the Laplacian operator. The high-frequency term restores reconstruction-suppressed detail; the small IQA regularizer limits amplification artifacts. No challenge images or quality labels are used in this step.

\subsection{Deterministic Inference Protocol}
Because many candidate gains are small, inference noise would obscure controlled comparisons. Sampling therefore uses a FlowMatch ODE with a fixed seed and CPU-side random-number generation, making repeated inference bit-exact within the evaluation environment. B1 uses guidance 2.5 and 14 steps (12.2\,s per view); B2 uses guidance 1.75 and 10 steps (about 9\,s per view). Both remain below the 30\,s limit.

\section{Experiments}
\label{sec:experiments}

\subsection{Evaluation Protocol and Reproducibility}
\label{sec:setup}
The official score combines normalized pixel fidelity, structural similarity, perceptual similarity, and no-reference image quality:
\[
S=\mathrm{PSNR}/40+\mathrm{SSIM}+(1{-}\mathrm{LPIPS})+\mathrm{NRIQA},
\]
where PSNR, SSIM, and LPIPS compare against ground truth~\cite{wang2004ssim,zhang2018lpips}, and NRIQA is the no-reference LAR-IQA score~\cite{avanaki2024lariqa}. Because a scalar total can hide opposing component changes, we report the complete score decomposition for every final submission.

Following Sec.~\ref{sec:methodology}, experiments distinguish three evidence levels. The early six-scene audit uses the official evaluator over 156 targets but supports only paired, within-lineage comparisons because its LoRA has seen all 40 scenes. The 34-scene sister lineage provides the scene-disjoint evidence used to select the final B2 recipe. Phase A then calibrates transfer through sparse paired submissions, whereas Phase B evaluates previously unseen scenes and determines the final ranking. All controlled inference comparisons use a fixed seed and deterministic ODE sampling.

As summarized in Table~\ref{tab:configs}, B1 and B2 train for 44.5\,h and 45\,h, respectively, on four A100-40G GPUs, with 21.4\,GB peak memory. Inference uses about 21\,GB on one A100-40G and takes 12.2\,s per view for B1 or about 9\,s for B2. Both remain well below the 72-hour training and 30-second inference limits. Repeated DDP training stays within the observed $\pm0.006$ band. Each isolated claim retains a paired before--after score under the same lineage and seed; only changes that exceed this band, replicate, or yield an expected transferred gain above 0.005 advance.

\begin{table}[t]
  \caption{Final submission configurations and measured resource profile.}
  \vspace{-1em}
  \label{tab:configs}
  \small
  \begin{tabular}{@{}p{1.85cm}p{2.75cm}p{2.75cm}@{}}
    \toprule
    \textbf{Setting} & \textbf{B1} & \textbf{B2} \\
    \midrule
    LoRA average & epochs 130/140/150 & epochs 120/130/140/150 \\
    Decoder data & 34-scene reconstruction & 40 scenes $+$ 11{,}273 open images \\
    Sampling & guidance 2.5, 14 steps & guidance 1.75, 10 steps \\
    Time per view & 12.2\,s & $\sim$9.0\,s \\
    Training time & 44.5\,h & 45.0\,h \\
    Peak memory & 21.4\,GB & 21.4\,GB \\
    \bottomrule
  \end{tabular}
\end{table}

\paragraph{Comparison scope.}
ZeroNVS, SV3D, MultiDiff, and CAT3D use cameras, geometry, joint generation, or training resources outside the challenge, so direct scores would violate the protocol. We instead compare exposure-matched, legal end-to-end variants; hidden Phase B measures unseen-scene performance.

\subsection{Final-Round Performance on Unseen Scenes}
\label{sec:results}
B1 is a conservative Phase-B baseline with three averaged checkpoints, a reconstruction-only decoder, and 14 sampling steps. After its largest unseen-scene losses appear in structural metrics, GoM selects three controlled changes for B2: a wider averaging window, high-frequency decoder refinement, and a shorter re-tuned inference protocol. Table~\ref{tab:phaseb} places the official leaderboard top three beside our B1 diagnostics. B1 is an internal submission from the same team, and its Phase-A row measures distribution shift rather than a competing final rank.

\begin{table}[t]
  \caption{Official Phase-B top three and GoM diagnostics. Bold marks the best component among ranked teams; diagnostic rows do not receive separate ranks.}
  \label{tab:phaseb}
  \small
  \setlength{\tabcolsep}{0.8pt}
  \begin{tabular}{@{}lrrrrrc@{}}
    \toprule
    \textbf{Entry} & \textbf{Total$\uparrow$} & \textbf{P/40$\uparrow$} & \textbf{SSIM$\uparrow$} & \textbf{$1{-}$LPIPS$\uparrow$} & \textbf{NRIQA$\uparrow$} & \textbf{Rank} \\
    \midrule
    GoM B2 (Ours) & \textbf{1.8628} & 0.3051 & \textbf{0.3282} & 0.4509 & \textbf{0.7786} & \shortstack{\textbf{1st}\\[-1pt]\textbf{Winner}} \\
    Rank-2 team & 1.8542 & 0.3013 & 0.3254 & \textbf{0.4698} & 0.7577 & 2nd \\
    Rank-3 team & 1.8488 & \textbf{0.3090} & 0.3197 & 0.4654 & 0.7547 & 3rd \\
    \midrule
    B1 (Phase B) & 1.8512 & 0.2998 & 0.3211 & 0.4557 & 0.7746 & --- \\
    B1 (Phase A) & 1.9386 & 0.3001 & 0.3749 & 0.4805 & 0.7831 & --- \\
    \bottomrule
  \end{tabular}
\end{table}

B1 loses 0.0874 when moving from overlapping Phase-A scenes to unseen Phase-B scenes. SSIM and $1{-}$LPIPS account for 0.0786, or about 90\%, of this gap, while normalized PSNR is essentially unchanged. The result directly supports the paper's central diagnosis: scene overlap hides structural and perceptual failures more than simple color error.

B2 raises the Phase-B total by 0.0116 and ranks first. The official top three are yyds (GoM, 1.8628), mg47717980aAc15 (1.8542), and mg3918280TPe23 (1.8488), giving GoM margins of 0.0086 and 0.0140. Relative to B1, the combined PSNR/SSIM gain is $0.0124$ and NRIQA adds $0.0040$, while a $0.0048$ loss in $1{-}$LPIPS offsets part of the gain. This matches the scene-disjoint diagnosis.

\subsection{Controlled Evidence for the Final Recipe}
Table~\ref{tab:gains} reports evidence for individual interventions. Each row belongs to its own controlled context; the deltas are evidence of repeatability, not additive contributions to the final score.

\begin{table}[t]
  \caption{Paired Phase-A evidence; rows are independently controlled and non-additive.}
  \label{tab:gains}
  \small
  \setlength{\tabcolsep}{4pt}
  \begin{tabular}{lrr}
    \toprule
    \textbf{Intervention} & \textbf{Paired online scores} & \textbf{$\Delta$} \\
    \midrule
    Seed selection & $1.8848\to1.8940$ & $+0.0092$ \\
    Checkpoint averaging & $1.8940\to1.9109$ & $+0.0169$ \\
    Adam restarts & $1.9136\to1.9245$ & $+0.0109$ \\
    Decoder adaptation (restart) & $1.9245\to1.9334$ & $+0.0089$ \\
    Decoder adaptation (continuous) & $1.9136\to1.9225$ & $+0.0089$ \\
    Three-point late averaging & $1.9245\to1.9314$ & $+0.0069$ \\
    \bottomrule
  \end{tabular}
\end{table}

\paragraph{Optimization trajectory.}
Under identical inference, restarts score 1.9245 versus 1.9136. Averaging yields the largest isolated Phase-A gain and transfer ratio (Table~\ref{tab:transfer}). Expanding from two to three late checkpoints raises the score from 1.9245 to 1.9314. The four-checkpoint overlap lineage reaches 1.9322, only $+0.0008$ beyond the three-point result and within the variation band. B2 is therefore selected by its scene-disjoint structural profile, not the largest overlapping-scene total.

\paragraph{Decoder bottleneck.}
Decoder adaptation raises reconstruction fidelity from 32.4 to 36.4\,dB and improves the scene-disjoint end-to-end score by 0.0114. Online, it adds exactly 0.0089 to both the restarted and continuous LoRA lineages (Table~\ref{tab:gains}), indicating that the decoder bottleneck is largely orthogonal to the DiT optimization trajectory. The same component adds 0.0072 to the final three-point lineage. This cross-lineage replication is stronger evidence than a single leaderboard delta and motivates adapting a measurable bottleneck before increasing model capacity.

\paragraph{Interaction of the two interventions.}
The interventions act at different stages. Restarts and averaging stabilize the conditional denoising trajectory and determine whether view-control gains survive a scene shift; decoder adaptation improves the RGB realization of an already formed latent. Their cross-lineage consistency therefore argues against a lucky checkpoint. It also explains why the gains are not mechanically additive: the decoder can restore detail and contrast, but cannot correct an upstream spatial hypothesis. B2 combines them because each addresses a measured failure mode under the same single-model inference budget.

\subsection{Memorization Gap and View Failures}
\label{sec:falsification}
A post-hoc structural audit localizes the remaining error upstream of image reconstruction. The adapted VAE reaches a reconstruction SSIM of 0.9765, whereas end-to-end synthesis on the held-out audit reaches only 0.2668. View direction is also consequential: left- and bottom-facing targets obtain approximately 0.17--0.21 SSIM, compared with 0.29--0.31 for top-facing views. The dominant limitation is therefore correspondence and disocclusion reasoning under large viewpoint changes, not decoder reconstruction fidelity.

Allowing a post-hoc global translation changes SSIM by only 0.0086, ruling out a simple coordinate-offset bug. The remaining discrepancy is spatially nonuniform and can remain visually plausible even when full-reference structural metrics fall.

Figure~\ref{fig:qualitative} complements the quantitative results with two target views per scene. Despite substantial changes in direction and scale, the final model preserves the dominant objects, scene appearance, and requested viewpoint across structured and natural environments. Performance is strongest when spatial anchors remain visible; newly exposed regions remain plausible but are less reference-aligned, consistent with the per-direction SSIM audit.

\subsection{Falsified Alternatives and Design Lessons}
The negative-result audit tests whether more capacity, more supervision, or more elaborate optimization can replace the final recipe. Table~\ref{tab:falsified} reports representative evidence, while the complete project ledger records more than 40 screened directions.

Two lessons emerge. First, gains come from trajectory control or components untouched by LoRA; simply enlarging or perturbing the backbone is ineffective. Second, weakening the one-to-one source--view mapping is especially harmful with 40 scenes. Because high-weight metric objectives cause structural regressions, GoM rejects gains seen only on overlapping scenes, unresolved ties, and metric-specific trade-offs. B2 instead uses a weak regularizer whose gain is jointly supported by PSNR and SSIM.

\begingroup
\setlength{\intextsep}{8pt plus 1pt minus 1pt}
\begin{table}[t]
  \caption{Representative falsifications; values are six-scene audit scores or deltas unless marked online.}
  \label{tab:falsified}
  \small
  \begin{tabular}{@{}p{1.65cm}p{5.90cm}@{}}
    \toprule
    \textbf{Family} & \textbf{Representative controlled evidence} \\
    \midrule
    \textbf{Capacity} & A 9B backbone reaches 1.8451 and rank-128 LoRA reaches 1.8055; neither exceeds the matched 4B rank-32 baseline. \\
    \textbf{Data/prompt} & Flips change the score by $-0.103$, all-pairs curricula by about $-0.11$, and prompt suffixes by $-0.056$. \\
    \textbf{Objectives} & High-weight evaluator-directed NRIQA losses reduce SSIM to 0.24--0.29; SGDR changes the score by $-0.14$ and DPO by $-0.046$~\cite{loshchilov2017sgdr,rafailov2023dpo}. \\
    \textbf{Composition} & Expert routing is $-0.0042$ online; TIES/DARE merging scores 1.6047/1.5856~\cite{yadav2023ties,yu2024dare}; encoder tuning lowers end-to-end performance by 0.019. \\
    \bottomrule
  \end{tabular}
\end{table}
\endgroup

\section{Discussion}

\subsection{Generalization-Aware Model Selection}
The first contribution is to make reliable model selection part of the method. Most novel-view synthesis work attributes progress to architecture, supervision, or geometric representation, yet our results show that the validation protocol can also change which model appears best. When source scenes overlap, a view-level split preserves scene identity and may reward late training or complex merging through memorization, making some improvements appear stronger than they are on unseen scenes. GoM therefore treats model selection as algorithmic rather than administrative by defining generalization through scene identity, requiring exposure-matched comparisons, separating overlapping diagnostics from scene-disjoint evidence, and calibrating offline improvements against online transfer.

This reframes validation as an interface between data design and optimization rather than a downstream reporting choice. The goal of model selection is not simply to maximize a convenient validation score, but to identify interventions that remain effective after a scene shift. The lesson extends beyond novel-view synthesis: whenever multiple outputs share the same underlying identity, such as camera views, video frames, or edits, output-level splits can overstate transformation ability by partially rewarding recall. Identity-disjoint selection should therefore be established before architecture or optimization choices are compared.

\subsection{Targeted End-to-End Adaptation}
GoM's second contribution is a compact adaptation recipe under the end-to-end constraint. Its key principle is stage-specific intervention: trajectory control improves generalization without enlarging the backbone, while decoder tuning targets image formation left untouched by LoRA. The failures of larger backbones, higher ranks, routing, and cross-lineage merging further suggest that, under data scarcity, measurable bottlenecks should be addressed before representational capacity is scaled.
End-to-end adaptation need not imply undifferentiated fine-tuning. A composite generator can remain a single inference model while its failure modes are localized and adapted separately. A practical strategy is therefore to identify whether error arises in conditional generation, reconstruction, or model selection, then apply the smallest intervention targeting that stage. GoM's gains with an unchanged backbone illustrate the data efficiency of this approach.

\subsection{Controlled Evidence and Design Lessons}
The third contribution is an evidence base built from 24 online submissions, more than 300 offline experiments, and over 40 falsified directions. Paired transfer calibration measures which offline gains survive online; cross-lineage decoder replications separate component effects from individual training trajectories; and negative results show that capacity, richer supervision, and metric-directed objectives are not substitutes for transferable view control. Together, these controls turn the final score into a set of testable design claims rather than an isolated competition result.

GoM also exposes an evaluation mismatch: hidden regions admit multiple plausible completions, whereas full-reference metrics reward only the recorded realization. The Phase-A-to-Phase-B gap and view-dependent SSIM ranges therefore motivate component-wise reporting, per-view distributions, and scene-level held-out splits. More generally, paired comparisons, transfer calibration, replication, and falsification provide mutually constraining evidence, making the resulting recipe more auditable and transferable than an unstructured collection of competition tricks.

\subsection{Scope and Limitations}
Our empirical findings should be interpreted within the limited-data setting of the challenge. Transfer ratios are estimated from a small number of paired submissions, and the six-scene audit, while useful for within-lineage model selection, is too small to support broad statistical conclusions. The current study is also restricted to a fixed set of discrete target views and does not evaluate continuous camera trajectories, multi-view consistency, or cross-dataset generalization. Future work should validate these findings on larger scene-disjoint benchmarks and extend GoM toward more general and consistent novel-view synthesis.

\section{Conclusion}

We presented GoM, the winning solution to the ACM MM 2026 Grand Challenge on Single-Image Guided Multi-Angle Image Synthesis, ranking first among 293 registered teams under a strict one-model, one-pass-per-view protocol. In the challenging 40-scene regime, we identify the key difficulty as learning transferable view control while avoiding scene memorization. GoM addresses this through scene-disjoint, exposure-matched model selection, trajectory regularization, and VAE decoder adaptation, enabling a geometry-free diffusion model to achieve strong structural and perceptual quality without test-time post-processing. Our results further suggest that, under limited data, reliable generalization-aware selection and targeted adaptation can be more effective than simply increasing model capacity.

\begin{acks}
\begin{sloppypar}
We thank Mango TV for organizing the Single-Image Guided Multi-Angle Image Synthesis Grand Challenge. This work is supported by the National Natural Science Foundation of China (No.~62402414), the Guangdong Basic and Applied Basic Research Foundation (No.~2025A1515011994), and Guangdong Provincial Project 2025D03J0014. It is also supported by Guangzhou Municipal Science and Technology Project (No.~2023A03J0011), the Guangzhou Industrial Information and Intelligent Key Laboratory Project (No.~2024A03J0628), and Guangdong Provincial Key Lab of Integrated Communication, Sensing and Computation for Ubiquitous Internet of Things (No.~2023B1212010007).
\end{sloppypar}
\end{acks}

\balance
\bibliographystyle{ACM-Reference-Format}
\bibliography{references}

@misc{flux2,
  author       = {{Black Forest Labs}},
  title        = {{FLUX.2-klein-4B}},
  year         = {2026},
  howpublished = {\url{https://huggingface.co/black-forest-labs/FLUX.2-klein-4B}},
  note         = {Model card, accessed August 22, 2026}
}

@inproceedings{lipman2023flow,
  author    = {Lipman, Yaron and Chen, Ricky T. Q. and Ben-Hamu, Heli and Nickel, Maximilian and Le, Matt},
  title     = {Flow Matching for Generative Modeling},
  booktitle = {International Conference on Learning Representations (ICLR)},
  year      = {2023}
}

@inproceedings{liu2023rectified,
  author    = {Liu, Xingchao and Gong, Chengyue and Liu, Qiang},
  title     = {Flow Straight and Fast: Learning to Generate and Transfer Data with Rectified Flow},
  booktitle = {International Conference on Learning Representations (ICLR)},
  year      = {2023}
}

@inproceedings{esser2024scaling,
  author    = {Esser, Patrick and Kulal, Sumith and Blattmann, Andreas and Entezari, Rahim and M{\"u}ller, Jonas and Saini, Harry and Levi, Yam and Lorenz, Dominik and Sauer, Axel and Boesel, Frederic and Podell, Dustin and Dockhorn, Tim and English, Zion and Lacey, Kyle and Goodwin, Alex and Marek, Yannik and Rombach, Robin},
  title     = {Scaling Rectified Flow Transformers for High-Resolution Image Synthesis},
  booktitle = {International Conference on Machine Learning (ICML)},
  year      = {2024}
}

@inproceedings{hu2022lora,
  author    = {Hu, Edward J. and Shen, Yelong and Wallis, Phillip and Allen-Zhu, Zeyuan and Li, Yuanzhi and Wang, Shean and Wang, Lu and Chen, Weizhu},
  title     = {{LoRA}: Low-Rank Adaptation of Large Language Models},
  booktitle = {International Conference on Learning Representations (ICLR)},
  year      = {2022}
}

@inproceedings{wortsman2022soups,
  author    = {Wortsman, Mitchell and Ilharco, Gabriel and Gadre, Samir Ya and Roelofs, Rebecca and Gontijo-Lopes, Raphael and Morcos, Ari S. and Namkoong, Hongseok and Farhadi, Ali and Carmon, Yair and Kornblith, Simon and Schmidt, Ludwig},
  title     = {Model Soups: Averaging Weights of Multiple Fine-Tuned Models Improves Accuracy Without Increasing Inference Time},
  booktitle = {International Conference on Machine Learning (ICML)},
  year      = {2022}
}

@inproceedings{izmailov2018swa,
  author    = {Izmailov, Pavel and Podoprikhin, Dmitrii and Garipov, Timur and Vetrov, Dmitry and Wilson, Andrew Gordon},
  title     = {Averaging Weights Leads to Wider Optima and Better Generalization},
  booktitle = {Conference on Uncertainty in Artificial Intelligence (UAI)},
  year      = {2018}
}

@inproceedings{huang2017snapshot,
  author    = {Huang, Gao and Li, Yixuan and Pleiss, Geoff and Liu, Zhuang and Hopcroft, John E. and Weinberger, Kilian Q.},
  title     = {Snapshot Ensembles: Train 1, Get {M} for Free},
  booktitle = {International Conference on Learning Representations (ICLR)},
  year      = {2017}
}

@inproceedings{loshchilov2017sgdr,
  author    = {Loshchilov, Ilya and Hutter, Frank},
  title     = {{SGDR}: Stochastic Gradient Descent with Warm Restarts},
  booktitle = {International Conference on Learning Representations (ICLR)},
  year      = {2017}
}

@inproceedings{rombach2022ldm,
  author    = {Rombach, Robin and Blattmann, Andreas and Lorenz, Dominik and Esser, Patrick and Ommer, Bj{\"o}rn},
  title     = {High-Resolution Image Synthesis with Latent Diffusion Models},
  booktitle = {IEEE/CVF Conference on Computer Vision and Pattern Recognition (CVPR)},
  year      = {2022}
}

@inproceedings{brooks2023instructpix2pix,
  author    = {Brooks, Tim and Holynski, Aleksander and Efros, Alexei A.},
  title     = {{InstructPix2Pix}: Learning to Follow Image Editing Instructions},
  booktitle = {IEEE/CVF Conference on Computer Vision and Pattern Recognition (CVPR)},
  year      = {2023}
}

@inproceedings{liu2023zero123,
  author    = {Liu, Ruoshi and Wu, Rundi and Van Hoorick, Basile and Tokmakov, Pavel and Zakharov, Sergey and Vondrick, Carl},
  title     = {Zero-1-to-3: Zero-Shot One Image to {3D} Object},
  booktitle = {IEEE/CVF International Conference on Computer Vision (ICCV)},
  pages     = {9264--9275},
  year      = {2023},
  doi       = {10.1109/ICCV51070.2023.00853}
}

@inproceedings{sargent2024zeronvs,
  author    = {Sargent, Kyle and Li, Zizhang and Shah, Tanmay and Herrmann, Charles and Yu, Hong-Xing and Zhang, Yunzhi and Chan, Eric Ryan and Lagun, Dmitry and Fei-Fei, Li and Sun, Deqing and Wu, Jiajun},
  title     = {{ZeroNVS}: Zero-Shot 360-Degree View Synthesis from a Single Image},
  booktitle = {IEEE/CVF Conference on Computer Vision and Pattern Recognition (CVPR)},
  pages     = {9420--9429},
  year      = {2024},
  doi       = {10.1109/CVPR52733.2024.00900}
}

@inproceedings{gao2024cat3d,
  author    = {Gao, Ruiqi and Holynski, Aleksander and Henzler, Philipp and Brussee, Arthur and Martin-Brualla, Ricardo and Srinivasan, Pratul P. and Barron, Jonathan T. and Poole, Ben},
  title     = {{CAT3D}: Create Anything in {3D} with Multi-View Diffusion Models},
  booktitle = {Advances in Neural Information Processing Systems 37 (NeurIPS)},
  pages     = {75468--75494},
  year      = {2024},
  doi       = {10.52202/079017-2403}
}

@inproceedings{voleti2024sv3d,
  author    = {Voleti, Vikram and Yao, Chun-Han and Boss, Mark and Letts, Adam and Pankratz, David and Tochilkin, Dmitry and Laforte, Christian and Rombach, Robin and Jampani, Varun},
  title     = {{SV3D}: Novel Multi-View Synthesis and {3D} Generation from a Single Image Using Latent Video Diffusion},
  booktitle = {European Conference on Computer Vision (ECCV)},
  pages     = {439--457},
  year      = {2024},
  doi       = {10.1007/978-3-031-73232-4_25}
}

@inproceedings{muller2024multidiff,
  author    = {M{\"u}ller, Norman and Schwarz, Katja and R{\"o}ssle, Barbara and Porzi, Lorenzo and Rota Bul{\`o}, Samuel and Nie{\ss}ner, Matthias and Kontschieder, Peter},
  title     = {{MultiDiff}: Consistent Novel View Synthesis from a Single Image},
  booktitle = {IEEE/CVF Conference on Computer Vision and Pattern Recognition (CVPR)},
  pages     = {10258--10268},
  year      = {2024},
  doi       = {10.1109/CVPR52733.2024.00977}
}

@article{wang2004ssim,
  author  = {Wang, Zhou and Bovik, Alan C. and Sheikh, Hamid R. and Simoncelli, Eero P.},
  title   = {Image Quality Assessment: From Error Visibility to Structural Similarity},
  journal = {IEEE Transactions on Image Processing},
  volume  = {13},
  number  = {4},
  pages   = {600--612},
  year    = {2004}
}

@inproceedings{zhang2018lpips,
  author    = {Zhang, Richard and Isola, Phillip and Efros, Alexei A. and Shechtman, Eli and Wang, Oliver},
  title     = {The Unreasonable Effectiveness of Deep Features as a Perceptual Metric},
  booktitle = {IEEE/CVF Conference on Computer Vision and Pattern Recognition (CVPR)},
  year      = {2018}
}

@inproceedings{avanaki2024lariqa,
  author    = {Jamshidi Avanaki, Nasim and Ghildyal, Abhijay and Barman, Nabajeet and Zadtootaghaj, Saman},
  title     = {{LAR-IQA}: A Lightweight, Accurate, and Robust No-Reference Image Quality Assessment Model},
  booktitle = {Computer Vision -- ECCV 2024 Workshops},
  pages     = {328--345},
  year      = {2025},
  doi       = {10.1007/978-3-031-91838-4_20}
}

@article{hosu2020koniq,
  author  = {Hosu, Vlad and Lin, Hanhe and Szir{\'a}nyi, Tam{\'a}s and Saupe, Dietmar},
  title   = {{KonIQ-10k}: An Ecologically Valid Database for Deep Learning of Blind Image Quality Assessment},
  journal = {IEEE Transactions on Image Processing},
  volume  = {29},
  pages   = {4041--4056},
  year    = {2020}
}

@inproceedings{hosu2024uhdiqa,
  author    = {Hosu, Vlad and Agnolucci, Lorenzo and Wiedemann, Oliver and Iso, Daisuke and Saupe, Dietmar},
  title     = {{UHD-IQA} Benchmark Database: Pushing the Boundaries of Blind Photo Quality Assessment},
  booktitle = {Computer Vision -- ECCV 2024 Workshops},
  pages     = {467--482},
  year      = {2025},
  doi       = {10.1007/978-3-031-91838-4_28}
}

@inproceedings{yadav2023ties,
  author    = {Yadav, Prateek and Tam, Derek and Choshen, Leshem and Raffel, Colin and Bansal, Mohit},
  title     = {{TIES}-Merging: Resolving Interference When Merging Models},
  booktitle = {Advances in Neural Information Processing Systems (NeurIPS)},
  year      = {2023}
}

@inproceedings{yu2024dare,
  author    = {Yu, Le and Yu, Bowen and Yu, Haiyang and Huang, Fei and Li, Yongbin},
  title     = {Language Models are Super Mario: Absorbing Abilities from Homologous Models as a Free Lunch},
  booktitle = {International Conference on Machine Learning (ICML)},
  year      = {2024}
}

@inproceedings{rafailov2023dpo,
  author    = {Rafailov, Rafael and Sharma, Archit and Mitchell, Eric and Manning, Christopher D. and Ermon, Stefano and Finn, Chelsea},
  title     = {Direct Preference Optimization: Your Language Model is Secretly a Reward Model},
  booktitle = {Advances in Neural Information Processing Systems (NeurIPS)},
  year      = {2023}
}

\end{document}